\documentclass[times, twoside, watermark]{zHenriquesLab-StyleBioRxiv}
\usepackage{blindtext} 
\usepackage{indentfirst}
\usepackage{graphicx}
\usepackage{makecell}
\usepackage{multirow}
\usepackage{enumitem}

\usepackage[explicit]{titlesec}
\titleformat{\section}
  {\large\rmfamily\bfseries}
  {\thesection}
  {0.5em}
  {#1}

\leadauthor{Bergs; Chung}

\renewcommand\Affilfont{\normalfont\small\itshape}
\renewcommand{\thesubsection}{\thesection.\arabic{subsection}}

\begin{document}

\title{ROS 2-Based LiDAR Perception Framework for Mobile Robots in Dynamic Production Environments, Utilizing Synthetic Data Generation, Transformation-Equivariant 3D Detection and Multi-Object Tracking}

\renewcommand\Authands{, }

\author[a,*]{Lukas~Bergs}
\author[a,b,*]{Tan~Chung}
\author[a,$\dagger$]{Marmik~Thakkar}
\author[a]{Alexander~Moriz}
\author[a]{Amon~Göppert}
\author[b]{Chinnawut~Nantabut}
\author[a,c]{Robert~Schmitt}

\affil[a]{Chair of Intelligence in Quality Sensing (IQS), Laboratory for Machine Tools and Production Engineering (WZL), RWTH Aachen University, Campus-Boulevard 30, Aachen 52074, Germany}
\affil[b]{The Sirindhorn International Thai-German Graduate School of Engineering (TGGS), King Mongkut’s University of Technology North Bangkok (KMUTNB), 1518 Pracharat 1 Road, Wongsawang, Bangsue, Bangkok 10800, Thailand}
\affil[c]{Fraunhofer Institute for Production Technology (IPT), Steinbachstr. 17, Aachen 52074, Germany}

\makeatletter
\def\@maketitle{%
  \newpage
  \begin{center}
    \vskip0.2em{\large\bfseries\@title\par}\vskip1.0em%
    {\lineskip.5em\large\@author\par}
     \vskip0.2em
    \begin{minipage}{\textwidth}
      \footnotesize
      \raggedright
      * Shared First Authorship; $\dagger$ Developed the synthetic data generation and model training infrastructure underlying the study’s results
    \end{minipage}
  \end{center}\par\vskip 0.5em
}
\makeatother

\maketitle

\begin{abstract}
Adaptive robots in dynamic production environments require robust perception capabilities, including 6D pose estimation and multi-object tracking. To address limitations in real-world data dependency, noise robustness, and spatiotemporal consistency, a LiDAR framework based on the Robot Operating System integrating a synthetic-data-trained Transformation-Equivariant 3D Detection with multi-object-tracking leveraging center poses is proposed. Validated across 72 scenarios with motion capture technology, overall results yield an Intersection over Union of 62.6\% for standalone pose estimation, rising to 83.12\% with multi-object-tracking integration. Our LiDAR-based framework achieves 91.12\% of Higher Order Tracking Accuracy, advancing robustness and versatility of LiDAR-based perception systems for industrial mobile manipulators.
\end {abstract}


\section{Introduction \& Structure}
Light Detection and Ranging (LiDAR) perception is a cornerstone of autonomous robotic systems, enabling situational awareness and informed decision-making in reconfigurable production environments. In industrial settings, mobile manipulators must reliably detect, localize, and interact with dynamic assets such as mobile workstations and mobile storage units. These capabilities rely fundamentally on 6D pose estimation, the determination of an object's 3D position and orientation, and multi-object tracking (MOT) to maintain spatial and temporal consistency of poses across successive observations. While 6D pose estimation enables precise robot docking to mobile workstations and accurate alignment with storage units for automated picking, real-time MOT ensures continuity by compensating for temporary occlusions, objects leaving the sensor’s field of view, motion during transport, and dynamic workspace changes. 

\begin{figure}[htbp]
    \centering
    \includegraphics[width=\linewidth]{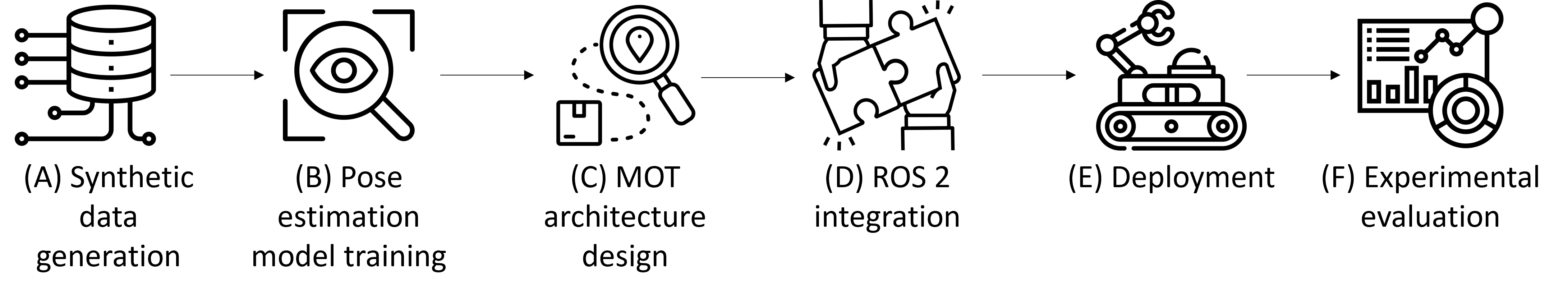}
    \captionsetup{skip=1pt}
    \caption{Framework overview. Icons by Freepik from www.flaticon.com.}
    \label{fig:framework}
\end{figure}

Despite advances in LiDAR-based 3D perception, several gaps hinder the deployment of reliable pose estimation and tracking systems in industrial scenarios. First, many learning-based pose estimation methods depend on large-scale real-world training datasets, which are labor-intensive to acquire and annotate. Second, LiDAR-based detection systems are susceptible to noise and occlusions, leading to pose inaccuracies that propagate to downstream tasks. Third, MOT frameworks often lack mechanisms to mitigate noise or adapt to rapid environmental changes, limiting their practicality in dynamic settings. To address these challenges, this work introduces a framework (see Fig.~\ref{fig:framework}) based on the Robot Operating System (ROS 2) \cite {ref1} unifying a LiDAR-based pose estimation and a MOT algorithm combining (1) a Transformation-Equivariant 3D Detection (TED) \cite {ref2} architecture trained with synthetic data to bypass real-data dependence; (2) an A Baseline for 3D Multi-Object Tracking and New Evaluation Metrics (AB3DMOT)-inspired \cite{ref3} tracker with center-pose based approach to achieve spatial and temporal pose consistency; and (3) a Design of Experiments (DoE) guided validation using motion capture technology across 72 scenarios in an industrial setting. By bridging sim-to-real gaps and integrating pose estimation with MOT, this work advances the robustness of LiDAR-based perception systems for industrial mobile manipulators. The results demonstrate improvements in pose accuracy and tracking performance, paving the way for reliable automation in production environments where rapid reconfiguration and operational efficiency are critical.

Structure: This paper is organized as follows. Section~\ref{sec:related_work} reviews related work in the field of synthetic data generation, 6D pose estimation and MOT. Section~\ref{sec:methodology} presents the methodology for a ROS 2-based framework along with the evaluation metrics. Section~\ref{sec:experiment_setup} details the experimental setup and the evaluation protocol. Section~\ref{sec:results} discusses the results of the experiments. Finally, Section~\ref{sec:conclusion} concludes the paper and outlines directions for future work.

\section{Related Work}
\label{sec:related_work}
\textbf{Synthetic data generation} has become essential for LiDAR-based perception, effectively mitigating the shortage of labeled real-world 3D point clouds \cite{ref4}. Leveraging high-fidelity 3D engines like NVIDIA Isaac Sim \cite{ref5} and Unreal Engine \cite{ref6}, simulation-based techniques generate ground truth annotated LiDAR data within virtual environments. Domain randomization techniques enhance generalization by varying environmental conditions during rendering \cite{ref7,ref8}, while  incorporating physically based sensor models and techniques such as ray tracing and path tracing to simulate real LiDAR behavior more accurately \cite{ref9}. Together, these approaches reduce the sim-to-real gap, streamlining training and validation for robust perception systems.

\textbf{LiDAR-based object pose estimation} algorithms determine an object's 6-degree-of-freedom pose (6-DoF), comprising its 3D position and orientation, from point cloud data, enabling applications like robotic manipulation \cite{ref10} and autonomous navigation \cite{ref11}. Classical methods have addressed these challenges through registration approaches, notably employing the Iterative Closest Point (ICP) algorithm and its variants to align observed point clouds with known object models; however, these techniques tend to be sensitive to initialization and computationally demanding \cite{ref12}. Recent deep learning approaches leverage point-based architectures such as PointNet \cite{ref13}, PointNet++ \cite{ref14}, and DGCNN \cite{ref15} to extract robust geometric features directly from raw data and predict 6-DoF poses using strategies based on keypoint detection or dense correspondence estimation. Recent studies further enhance detection performance by integrating multi-scale attention mechanisms as demonstrated by CasA \cite{ref16} and by using image guidance to refine spatial context in depth completion \cite{ref17}. More recently, TED incorporates specialized layers (TeSpConv, TeBEV pooling, TiVoxel pooling) to ensure consistent feature responses under spatial transformations \cite{ref2}.

\textbf{MOT} has evolved significantly from early Tracking-by-Detection frameworks like AB3DMOT \cite{ref3}, which relied on basic matching based on Intersection over Union (IoU) with the Hungarian Algorithm. Recent approaches improve association accuracy by incorporating richer features such as 3D appearance and motion coherence \cite{ref18,ref19,ref20}, improvement is also achieved through capturing uncertainties in position, size, and orientation by using multi-dimensional Gaussian distributions and Kullback-Leibler divergence \cite{ref21}. Recent prediction modules replace Kalman Filtering with advanced techniques like Adaptive Cubature Kalman Filters \cite{ref22}, Constant Acceleration model \cite{ref23}, and Constant Acceleration with Angular Velocity model  \cite{ref24}, or geometry-aware optimizers such as ghost-aware trajectory postoptimizer in object-aware anti-occlusion 3D MOT \cite{ref20}. Although complex and computationally expensive approaches, such as inter-frame bipartite graphs constructed from appearance and spatial features, enhance tracking quality \cite{ref25}, lightweight center-point-based approaches achieve competitive robustness while significantly reducing computational cost \cite{ref26}, aligning with our focus on maintaining a lightweight design.

\section{Methodology}
\label{sec:methodology}

\begin{figure}[htbp]
    \centering
    \includegraphics[width=0.8\linewidth]{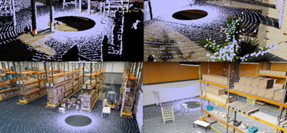}
    \captionsetup{skip=1pt}
    \caption{Synthetic data generation visualization.}
    \label{fig:synthetic_data}
\end{figure}

\textbf{Synthetic data generation}. Our data generation pipeline leverages NVIDIA Isaac Sim's physics-based simulation platform, built on Omniverse, to generate synthetic LiDAR datasets for 6-DoF pose estimation training. The pipeline utilizes Omniverse Replicator for domain randomization and realistic sensor modeling, configuring an RTX-based LiDAR via JSON parameters, defining intrinsic parameters (e.g., resolution, range, beam divergence) and realistic noise profiles to accurately emulate an Ouster OS1-128 sensor. To ensure diversity and enhance real-world generalization, the pipeline dynamically loads varied simulated environments, randomizes object layouts and LiDAR viewpoints. Additionally, it applies object-level augmentations, including scaling (±15\%), full 360° yaw rotations, and varying instance counts (1–5) to further prevent overfitting and enrich the training data. Ground truth labels follow the OpenPCDet standard format, with a ray-hit filter ensuring only LiDAR-visible objects are annotated \cite{ref27}. A dataset comprising 15,000 samples ($\approx$ 26.1 GB) was generated, including all object categories. Separate models were trained for each category using a 50\% train-test split. Fig.~\ref{fig:synthetic_data} illustrates the synthetic data generation, highlighting the diversity of simulated environments and the application of object-level augmentation.

\begin{figure}[htbp]
    \centering
    \includegraphics[width=\linewidth]{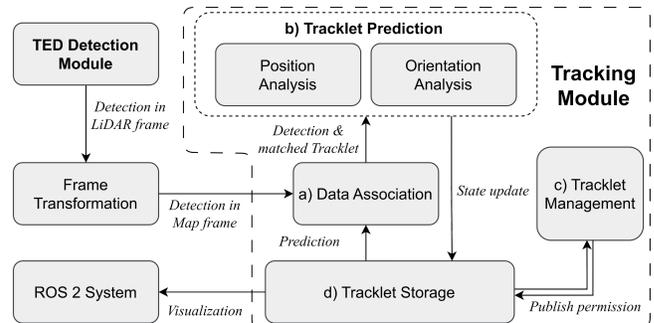}
    \captionsetup{skip=1pt}
    \caption{Architecture of MOT Module.}
    \label{fig:MOT_module}
\end{figure}

\textbf{Pose estimation module}. Fig.~\ref{fig:MOT_module} presents the architecture of the implemented detection and MOT algorithm.  The TED network implementation is based on the authors' original code, modifying only the data pipeline to accommodate synthetic data inputs while preserving all architectural parameters, such as network architecture, learning rate, optimizer settings, and batch size \cite{ref2}. This controlled approach isolates the impact of synthetic data augmentation, enabling clear evaluation of its contribution to model generalization. Changes in performance can therefore be directly attributed to the quality and diversity of our synthetic training data set.

\textbf{Tracking module} comprises four key modules: \textit{(a) Data Association, (b) Tracklet Prediction, (c) Tracklet Management,} and \textit{(d) Tracklet Storage}.

\textit{a) Data Association:} After transforming the detection from LiDAR frame to global (map) frame. This module employs a lightweight, center-based approach, utilizing a greedy algorithm that calculates Euclidean distances between detected object centers and existing tracklets \cite{ref26}. A new tracklet is initiated only if the minimum distance exceeds a threshold based on the objects’ sizes, ensuring efficient association without reliance on computationally intensive appearance or geometric features.

\textit{b) Tracklet Prediction:} Designed to stabilize detection noise, this module comprises two subcomponents: Position Analysis and Orientation Analysis. For stationary objects, it averages historical positions and orientations to mitigate fluctuations. Orientation outliers are identified by comparing incoming detections with historical data; if discrepancies exceed a threshold, the orientation history is updated to reflect current observations. For symmetric objects prone to orientation flipping, the module generates multiple orientation hypotheses based on the quantity of symmetry planes and selects the one closest to the previous orientation, ensuring continuity.  When movement is detected based on positional (> 0.05 m) and rotational (> 2.5°) differences between consecutive poses, the module transitions to using real-time detection data to accurately track dynamic changes.

\textit{c) Tracklet Management:} This component filters out unreliable tracklets by analyzing the consistency and duration (3 detections in 2 s) of detection matches over time. Only tracklets with sustained, stable associations are retained, enhancing the reliability of the tracking system.

\begin{figure}[htbp]
    \centering
    \includegraphics[width=\linewidth]{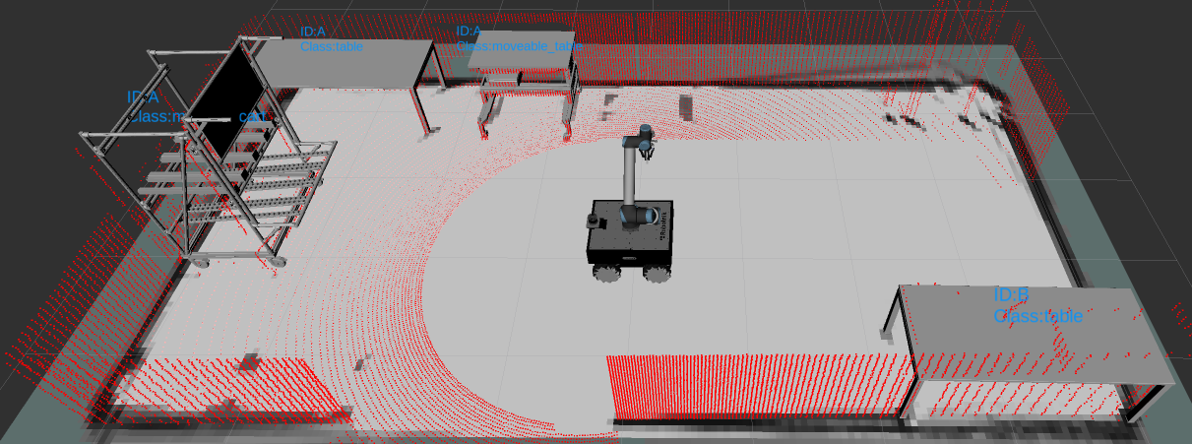}
    \captionsetup{skip=1pt}
    \caption{Tracklets visualization in RViz.}
    \label{fig:tracklet_visualization}
\end{figure}

\textit{d) Tracklet Storage:} Serving as centralized storage, this module maintains comprehensive information on all active tracklets. It facilitates data access and updates across the system, broadcasting tracked transforms and publishing mesh markers to ROS 2 for visualization in RViz as shown in Fig.~\ref{fig:tracklet_visualization}. This integration enables real-time rendering of 3D object models, enhancing system transparency.

\textbf{Evaluation metrics.} To assess the framework's performance in both standalone 6D pose estimation and integrated MOT, spatial accuracy is measured using IoU and Root Mean Square Error (RMSE) between predicted and ground truth bounding boxes, indicating how precise objects are localized. To evaluate the tracking capabilities of the system, Detection Accuracy (DetA) and Higher Order Tracking Accuracy (HOTA) are employed \cite{ref28}. DetA quantifies the system's ability to correctly detect objects, independent from their identity across frames. It is computed as the ratio of true positive detections to the sum of true positives, false positives, and false negatives, capturing the balance between missed and false detections. For each frame, a predicted object is counted as a true positive (TP) if it overlaps with a ground-truth object (IoU > 0.5), as a false positive (FP) if no overlap is found, and as a false negative (FN) if a ground-truth object has no matching prediction. HOTA evaluates both detection and association accuracy, providing a balanced measure of tracking performance by considering the trade-off between detecting objects and maintaining consistent identities across time.

\section{Experiment Setup}
\label{sec:experiment_setup}
\begin{figure}[htbp]
    \centering
    \includegraphics[width=\linewidth]{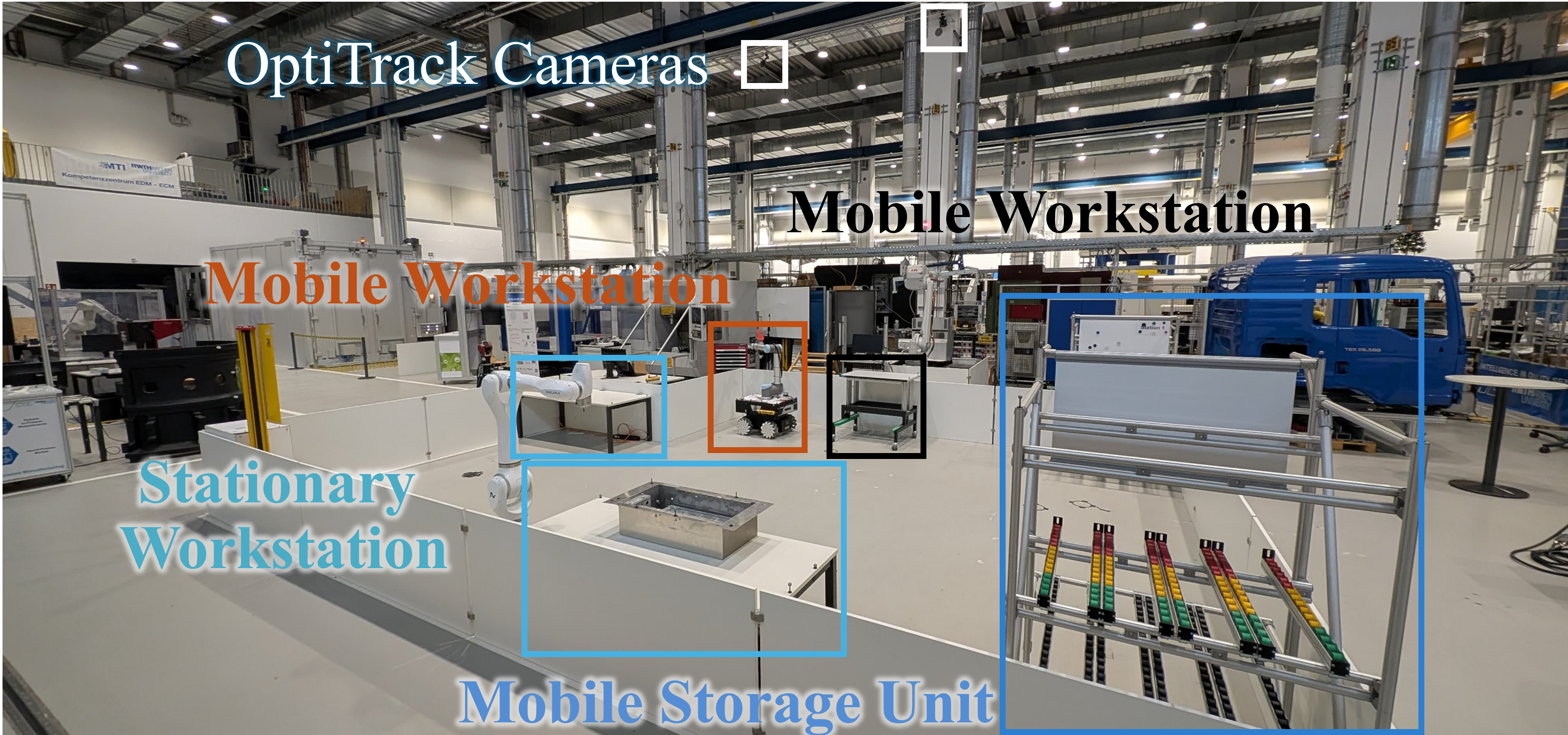}
    \captionsetup{skip=1pt}
    \caption{Experimental environment at WZL-IQS.}
    \label{fig:experimental_environment}
\end{figure}

Experimental trials evaluate all key factors from Table~\ref{tab:experimental_factors} at their specified discrete levels. Therefore, the Mixed-Level Orthogonal Array (OA) DoE, with rows as tests and columns as mix-level factors, is employed to evaluate our LiDAR-based 6D pose estimation and MOT framework across 72 systematically structured trials \cite{ref29}. Fig.~\ref{fig:experimental_environment} shows the 44 m² reconfigurable industrial-like lab production environment, containing the mobile manipulator, stationary and mobile workstations, mobile storage units, the OptiTrack motion capture system, as well as dynamically placed occlusion objects used for experimental validation. This controlled setup enables isolation of performance factors while maintaining validity through industrial-relevant layouts. To accommodate these mixed-level factors, the OA matrix from Table~\ref{tab:OA_matrix} is applied for the mobile workstation and mobile storage unit \cite{ref30}. Full experimental details are hosted on Zenodo \cite{ref31}. 

The mobile manipulator platform used is the RB-Kairos, which combines an omnidirectional wheelbase and an UR10 robotic arm. The robot is equipped with a sensor suite, including an Inertial Measurement Unit (IMU), Hall effect sensors and dual SICK S300 laser scanners for SLAM-based localization and an Ouster OS1-128 3D LiDAR for 6D pose estimation. To ensure accurate ground truth data for both the robot and surrounding detectable assets, the OptiTrack photogrammetry system is employed. The motion capture setup, comprising 38 Primex 41 cameras, delivers millimeter-level tracking precision \cite{ref32}. Reflective markers are affixed to robot and detectable assets, with their positions measured using a coordinate measuring machine. XML files are generated from the measured coordinates, defining the rigid bodies, with origins adjusted to align with either a known robot reference frame or the bounding box centers of detectable objects.

\begin{table}[htbp]
    \centering
    \caption{Levels of input factors (PL: with object linear movement, PA: with object angular movement, NL: without object linear movement, NA: without object angular movement).}
    \label{tab:experimental_factors}
    \fontsize{7}{10}\selectfont
    \resizebox{0.48\textwidth}{!}{%
    \begin{tabular}{lccc}
        \hline
        \noalign{\vskip 1pt}
        Input Factors & Level 1 & Level 2 & Level 3 \\
        \hline
        Number of Objects & 1 & 2 &  \\
        Robot Linear Movement & Stationary & 0.25 m/s & 0.5 m/s \\
        Robot Angular Movement & Stationary & 0.25 rad/s & 0.5 rad/s \\
        Occlusion & None & $< 20\%$ & $> 40\%$ \\
        Initial Distance & 2.5 m & 3.5 m & 4.5 m \\
        \hline
        \multirow[t]{2}{5.2cm}[0.6em]{Robot Linear Movement\\with Objects Movements\\(for mobile objects only,\\6 levels)}
        & \makecell[l]{Stationary\\- NL - NA}
        & \makecell[l]{Stationary\\- PL - NA}
        & \makecell[l]{Stationary\\- NL - PA} \\
        & \makecell[l]{Stationary\\- PL - PA}
        & 0.25 m/s
        & 0.5 m/s \\
        \hline
    \end{tabular}
    }
\end{table}

The robotic system's communication and software framework is built upon ROS 2, providing the middleware and tools for communication and coordination between the software components deployed on the robot's onboard computer and the application server. Wireless communication is facilitated through a 5G network and zenoh middleware, ensuring high-speed, low-latency data transmission \cite{ref33}. On the robot’s onboard hardware, motion execution is controlled by the control software, which interprets high-level commands and converts them into motor commands. Sensor drivers collect data from various sensors, including laser scanners, 3D LiDAR, IMU, and rotary encoders. To optimize bandwidth usage, data filtering and compression techniques are applied, mainly for transmitting point cloud data between the robot and the application server. The application server serves as the central processing hub, hosting ROS 2 modules for localization, 6D pose estimation MOT and visualization as well as the client for streaming motion capture data from the OptiTrack system. The application server is equipped with dual AMD EPYC 75F3 processors (32 cores, 2.95 GHz), 256 GB of RDIMM memory (16×16 GB, 3200 MT/s), and an NVIDIA Ampere A10 GPU with 24 GB of memory to support real-time computation and data processing.

\begin{table}[htbp]
    \centering
    \caption{OA matrix for mobile workstation and mobile storage units (Single mobile workstation: 1 - 18; Single mobile storage unit: 1 - 18; Two mobile storage units: 19 - 36).}
    \label{tab:OA_matrix}
    \fontsize{7}{10}\selectfont
    \resizebox{0.48\textwidth}{!}{%
    \begin{tabular}{llllc}
        \hline
        \noalign{\vskip 1pt}
        \makecell[t]{Number} & \makecell[t]{Robot Linear} & \makecell[t]{Robot\\Angular} & \makecell[t]{Occlusion} & \makecell[t]{Initial\\Distance} \\
        \hline
        1--19  & Stationary - NL - NA & Stationary    & No        & 2.5 m \\
        2--20  & Stationary - NL - NA & 0.25 rad/s   & $< 20\%$  & 3.5 m \\
        3--21  & Stationary - NL - NA & 0.5 rad/s    & $> 40\%$  & 4.5 m \\
        4--22  & Stationary - PL - NA & Stationary    & No        & 3.5 m \\
        5--23  & Stationary - PL - NA & 0.25 rad/s   & $< 20\%$  & 4.5 m \\
        6--24  & Stationary - PL - NA & 0.5 rad/s    & $> 40\%$  & 2.5 m \\
        7--25  & Stationary - NL - PA & Stationary    & $< 20\%$  & 2.5 m \\
        8--26  & Stationary - NL - PA & 0.25 rad/s   & $> 40\%$  & 3.5 m \\
        9--27  & Stationary - NL - PA & 0.5 rad/s    & No        & 4.5 m \\
        10--28 & Stationary - PL - PA & Stationary    & $> 40\%$  & 4.5 m \\
        11--29 & Stationary - PL - PA & 0.25 rad/s   & No        & 2.5 m \\
        12--30 & Stationary - PL - PA & 0.5 rad/s    & $< 20\%$  & 3.5 m \\
        13--31 & 0.25 m/s             & Stationary    & $< 20\%$  & 4.5 m \\
        14--32 & 0.25 m/s             & 0.25 rad/s   & $> 40\%$  & 2.5 m \\
        15--33 & 0.25 m/s             & 0.5 rad/s    & No        & 3.5 m \\
        16--34 & 0.5 m/s              & Stationary    & $> 40\%$  & 3.5 m \\
        17--35 & 0.5 m/s              & 0.25 rad/s   & No        & 4.5 m \\
        18--36 & 0.5 m/s              & 0.5 rad/s    & $< 20\%$  & 2.5 m \\
        \hline
    \end{tabular}
    }
\end{table}

\section{Results}
\label{sec:results}

\subsection{Quantitative Analysis}

\begin{table}[htbp]
    \centering
    \caption{Overall experiment results. MW - mobile workstation; SW – station-ary workstation; MSU - mobile storage unit; Ave – Average of three classes, D - Detection; T - Tracklet. IoU (Average IoU); Pos (Directional RMSE) Rot (Yaw RMSE). Full experimental details are provided in the \cite{ref31}.}
    \label{tab:results}
    \fontsize{7}{10}\selectfont
    \resizebox{0.48\textwidth}{!}{%
    \begin{tabular}{c|c c c c c c}
        \hline
        & & IoU & Pos & Rot & DetA & HOTA \\
        \hline
        \multirow{2}{*}{MW}
            & D & 40.03 \% & 1.78 $m$ & 68.99$^\circ$ & 49.38 \% & - \\
            & T & 79.08 \% & 0.06 $m$ & 23.11$^\circ$ & 76.80 \% & 87.31 \% \\
        \hline
        \multirow{2}{*}{SW}
            & D & 72.20 \% & 0.71 $m$ & 102.49$^\circ$ & 66.80 \% & - \\
            & T & 82.23 \% & 0.05 $m$ & 1.23$^\circ$ & 85.06 \% & 92.11 \% \\
        \hline
        \multirow{2}{*}{MSU}
            & D & 75.79 \% & 1.14 $m$ & 29.52$^\circ$ & 89.57 \% & - \\
            & T & 88.04 \% & 0.04 $m$ & 8.63$^\circ$ & 90.98 \% & 93.92 \% \\
        \hline
        \multirow{2}{*}{Ave}
            & D & 62.67 \% & 1.21 $m$ & 67.00$^\circ$ & 68.58 \% & - \\
            & T & 83.12 \% & 0.05 $m$ & 10.99$^\circ$ & 84.28 \% & 91.12 \% \\
        \hline
    \end{tabular}
    }
\end{table}

\textit{a) IoU:} According to Table~\ref{tab:results}, the integrated MOT system achieved an average IoU of 83.12\% across all object classes and experiments, representing a 20.45\% improvement over the standalone detection system (62.67\%). The class MSU showed particularly strong performance with an averaged IoU for both units of 93.92\% under optimal conditions (stationary, no occlusion, close distance), while challenging scenarios (high speed, > 40\% occlusion, far distance) maintained an IoU of 86.75\% with MOT.  

\textit{b) Directional RMSE (position):} The MOT module reduced positional errors from 1.21 m to 0.05 m on average. For the simplest case (stationary, no occlusion, close distance), centimeter-level precision (0.0182 m RMSE) for the class MW is noticed. The most significant improvement for MW was observed in the experiment characterized by low robot movement, high occlusion, and a medium object distance. In this case, the RMSE decreased from 2.621 m to 0.157 m.
Table 3. Overall experiment results. MW - mobile workstation; SW – station-ary workstation; MSU - mobile storage unit; Ave – Average of three classes, D - Detection; T - Tracklet. IoU (Average IoU); Pos (Directional RMSE) Rot (Yaw RMSE). Full experimental details are provided in the \cite{ref31}.

\textit{c) Yaw RMSE (yaw rotation):} Similarly, the experiments revealed improvement of average orientation estimation for all classes from 67° to 10.99° RMSE with MOT integration. The orientation continuity check successfully resolved symmetric object ambiguities, reducing flipping errors from 138.6° to 0.64° in the worst-case scenario for the class SW.

\textit{d) DetA:} The MOT-enhanced system maintained 84.28\% average DetA (68.58\% for detection-only), with the class MSU achieving 90.98\% in optimal conditions. Even in challenging high-speed, high-occlusion scenarios, DetA for a single MSU remained above 83\%.

\textit{e) HOTA:} The system achieved an average HOTA score of 91.12\% across all classes, with the class MSU performing best with an average HOTA score of 93.92\%. Only one ID switch occurred across all trials for the class MSU, resulting in the HOTA score of 67.57\%.

\subsection{Qualitative Observations}

\begin{figure}[htbp]
    \centering
    \includegraphics[width=\linewidth]{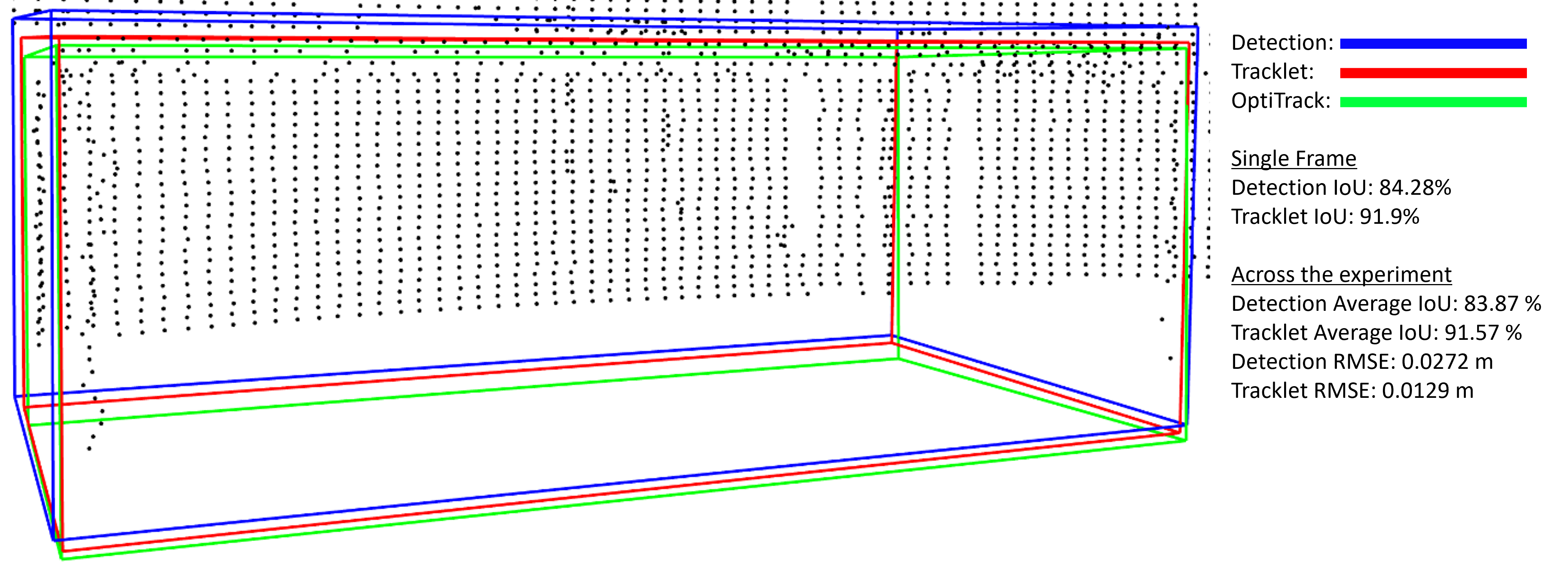}
    \captionsetup{skip=1pt}
    \caption{Bounding box overlap of SW experiment.}
    \label{fig:bounding_box}
\end{figure}

The experimental evaluation across 72 trials revealed several key insights into the system's performance. MSU achieved superior performance (93.92\% HOTA) compared to the other classes due to its distinctive 1.8 m profile and unambiguous point cloud features. Conversely, MW (0.7 m) and SW (0.82 m) exhibit higher false positive rates, as their lower profiles and simpler geometries increase confusion with similarly shaped background structures, underscoring how object geometry influences detection quality.

\begin{figure}[htbp]
    \centering
    \includegraphics[width=\linewidth]{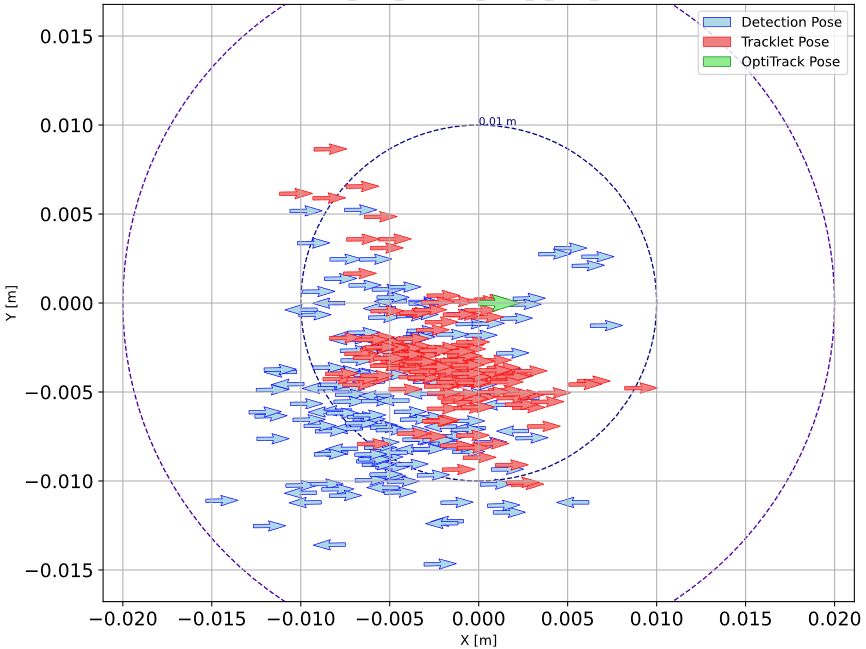}
    \captionsetup{skip=1pt}
    \caption{Directional and rotation errors in SW experiment.}
    \label{fig:errors}
\end{figure}

Compared to the detection-only configuration the MOT-enhanced approach demonstrated consistent pose estimation accuracy improvements across all object classes. Most notably, the system reduced yaw rotation errors for the symmetric class SW by 98.8\% (102.5° to 1.23° RMSE) through temporal consistency checks and symmetry-aware filtering. Fig.~\ref{fig:bounding_box} and Fig.~\ref{fig:errors} visualize the improvement of IoU and pose similarity of tracklet over detection. This improvement came with an inherent initialization tradeoff: the requirement of 3-5 consecutive detections ($\approx$2 s) introduces temporary latency but ultimately enhanced accuracy through robust false positive rejection, reducing overall RMSE despite slightly compromising initial DetA in static scenarios. 

The analysis further identified robot rotation as a critical performance factor. During rotational movement, delays in point cloud transferring to the application server create temporal misalignment between sensor data and robot state. This asynchrony produces two distinct error modes: (1) apparent point cloud distortion proportional to angular velocity, and (2) incorrect ego-motion compensation during coordinate transformation. The resulting pose errors grows quadratically with both angular velocity and latency, becoming particularly pronounced during high-speed rotations.

System performance demonstrated consistent degradation patterns under dynamic conditions, with both increasing object velocities and occlusion levels correlating with larger positional and angular offsets. These findings highlight that while the MOT system provides significant improvements in pose stability and temporal consistency, its effectiveness remains fundamentally dependent on the reliability of the underlying pose estimation algorithm, particularly in challenging scenarios.

\section{Conclusion \& Future Work}
\label{sec:conclusion}

This study presents a robust LiDAR-based 6D pose estimation and tracking system for mobile robots within flexible industrial settings, demonstrating three key advances: (1) a synthetic-data-trained pose estimation model that eliminates reliance on real-world datasets, (2) a tracking framework that improves noise robustness and spatiotemporal consistency, and (3) a quantitative assessment and characterization of performance factors through a DoE-based evaluation. Experimental results demonstrate an IoU of 62.67\% for standalone pose estimation, which increases to 83.12\% when integrated with MOT, alongside a HOTA score of 91.12\%. Centimeter-level accuracy in static scenarios (< 4 cm RMSE directional) and robust dynamic tracking were achieved, enabling reliable robotic interaction with mobile assembly and logistics resources. The perception system enables mobile manipulators to safely approach dynamic workstations and storage units. While residual errors may occur, these are compensable by the manipulator's closed-loop control, facilitating precision tasks like part picking and assembly in unstructured environments. 

Future efforts will focus on addressing the identified limitations and expanding the framework’s applicability:
\begin{itemize}[itemsep=2pt, topsep=2pt, parsep=0pt, partopsep=0pt]{
    \item Latency optimization: Investigate and mitigate delays between point cloud acquisition, pose estimation, and robot localization to improve performance in high-velocity dynamic scenarios.
    \item Multi-modal perception: Integrate RGB-D camera for fine-grained pose estimation of workpieces located on approached stations, enabling precise manipulation post-alignment.
    \item Digital twin integration: Implement persistent asset tracking by coupling with a digital twin system to log pose histories and object metadata in dynamic environments.
    \item Cross-domain deployment: Adapt the framework for applications in the automotive domain, where comparable challenges in dynamic object perception, tracking, and environmental understanding are prevalent.}
\end{itemize}

\section*{Acknowledgements}
This work is part of the research project KaliMoRo and funded by the German Federal Ministry for Economic Affairs and Climate Action (BMWK) through the Industrial Collective Research (IGF) program.

\section*{Supplementary Material}
The datasets collected from the experiments, along with all
results and scripts used to evaluate the data, are availably accessible on Zenodo \cite{ref31}.

\end{document}